\documentclass[runningheads]{llncs}
\usepackage{graphicx}

\usepackage{amsmath,amssymb} 
\usepackage{color}

\usepackage{multirow}
\usepackage[T2A,T1]{fontenc}

\usepackage{subfigure}

{\raise0.5ex\hbox{$#1$}\! \left/ \! \lower0.5ex\hbox{$#2$}\right.}

\begin{document}

\title{An unsupervised deep learning framework via integrated optimization of representation learning and GMM-based modeling} 
\titlerunning{Unsupervised Deep GMM Modeling} 


\author{Jinghua Wang\orcidID{0000-0002-2629-1198} \and
Jianmin Jiang\orcidID{0000-0002-7576-3999}\thanks{Corresponding author}}
%

\authorrunning{J. Wang and J. Jiang} 


\institute{Research Institute for Future Media Computing,College of Computer Science \& Software Engineering, Shenzhen University, Shenzhen, China\\
\email{wang.jh@szu.edu.cn, jianmin.jiang@szu.edu.cn}}

\maketitle

\begin{abstract}
While supervised deep learning has achieved great success in a range of applications, relatively little work has studied the discovery of knowledge from unlabeled data. In this paper, we propose an unsupervised deep learning framework to provide a potential solution for the problem that existing deep learning techniques require large labeled data sets for completing the training process. Our proposed introduces a new principle of joint learning on both deep representations and
GMM (Gaussian Mixture Model)-based deep modeling, and thus an integrated objective function is proposed to facilitate the principle. 
In comparison with the existing work in similar areas, our objective function has two learning targets, which are created to be jointly optimized to achieve the best possible unsupervised learning and knowledge discovery from unlabeled data sets. While maximizing the first target enables the GMM to achieve the best possible modeling of the data representations and each Gaussian component corresponds to a compact cluster, maximizing the second term will enhance the separability of the Gaussian components and hence the inter-cluster distances. As a result, the compactness of clusters is significantly enhanced by reducing the intra-cluster distances, and the separability is improved by increasing the inter-cluster distances. Extensive experimental results show that the propose method can improve the clustering performance compared with benchmark methods.

\end{abstract}

\section{Introduction}

With the advanced machine learning technologies, we can process the explosion data effectively. Deep learning is one of the most popular techniques, and has been successfully applied in many computer vision tasks, such as image classification \cite{He2016DeepResidue,spn-tcsvt}, semantic segmentation  \cite{Noh2015Learning,wang-eccv2016}, and object detection \cite{Ren2015Faster}. However, these techniques  \cite{He2016DeepResidue,Noh2015Learning,Ren2015Faster,wang-eccv2016} heavily rely on a huge number of high quality labeled training data to learn a good model. Yet, manually labeling the training data is extremely time-consuming. Thus, it is necessary to develop unsupervised techniques that can discover knowledge from the easily available unlabeled data.

Clustering is one of the most popular unsupervised machine learning techniques.
Traditional clustering methods, such as k-means \cite{Lloyd1982Least} and Gaussian Mixture Model (GMM) \cite{Bishop:2006:PRM:1162264}, categorize samples by investigating their similarities directly in the original data space. Thus, their performances heavily depend on the distribution of the data samples \cite{Aggarwal:2013:DCA:2535015}. 


In order to achieve robustness against the data sample distributions, researchers propose to extract features or learn representations before conducting the clustering procedure \cite{Hinton-science-2006,you-2016-cvpr-scalable-sparse-subspace}. While the supervised methods learn representations which are closely correlated with the class labels, unsupervised representation learning is more difficult due to the unavailability of label information. Through representation learning, the methods  \cite{Hinton-science-2006,you-2016-cvpr-scalable-sparse-subspace} can explicitly or implicitly discover the hidden variables which are more discriminative than the original data sample.
While the methods \cite{Hinton-science-2006,you-2016-cvpr-scalable-sparse-subspace} successfully learn discriminative representations for various tasks, the resulting representations are not necessarily the optimal choice for the clustering task. 
To learn representations of data samples that are catered for the clustering task,  researchers  \cite{yangCVPR2016joint,yang_icml2017-pmlr-v70-yang17b,Xie-icml-2016} propose to integrate the representation learning with clustering.

\begin{figure*}[t]
	\begin{center}
		\includegraphics[width=0.9\linewidth]{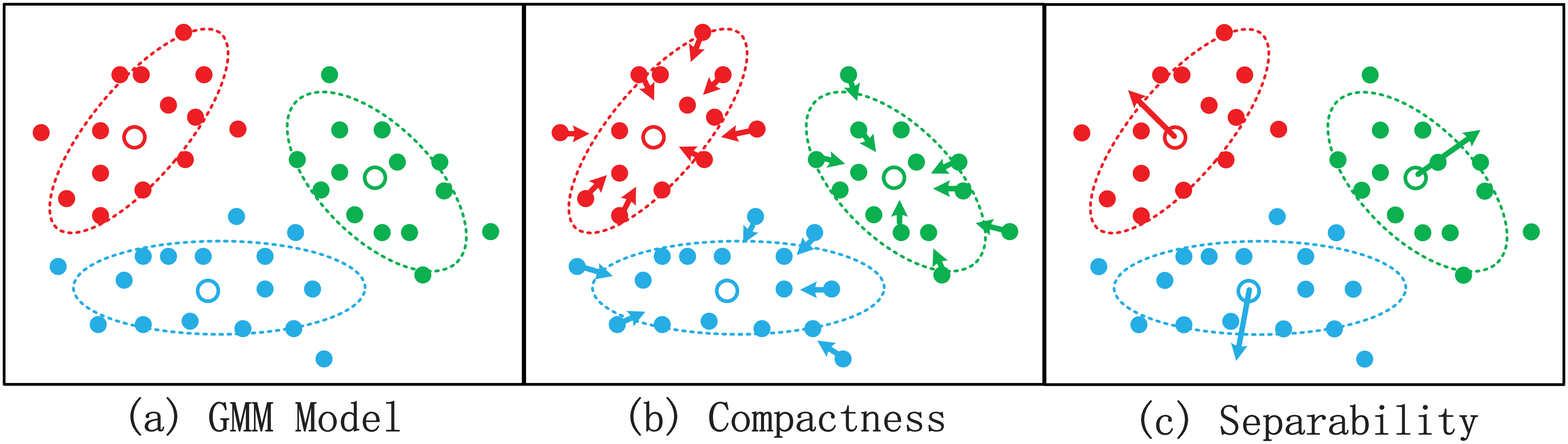}
	\end{center}
	\caption{GMM modeling (best viewed in color). (a) Traditional GMM method fits the data points by a set of Gaussian components. (b) Adjust the sample representations towards the Gaussian center to improve the compactness. (c) Adjust the Gaussian centers to enhance the separability between different components. }
	\label{fig:GaussianDistributions}
\end{figure*}

In this paper, we propose a new joint optimization approach for unsupervised representation learning and clustering. We aim at formulating a GMM out of the whole data representations and the center of each Gaussian component represents a cluster center. 
Modeling the representations by a GMM significantly alleviates the constraints in the work  \cite{Stuhlsatz_feature_extraction_TNNLS2012} and  \cite{Wang2017-cvpr-oral-ggdenet}, which model the representations by a single Gaussian model. Our approach aims to learn data representations which are intra-cluster compact and inter-cluster separable.

As in all of the GMM-based methods, we maximize the GMM likelihood to discover a feasible GMM for the whole data representations, as shown in Fig. \ref{fig:GaussianDistributions} (a). An important but rarely mentioned point in a GMM model is: a larger GMM likelihood also means a smaller distance between a sample representation and its associating cluster center (in addition to a set of well positioned Gaussian centers). In our approach, both the cluster center and the representation are iteratively updated in the training procedure.  Thus, by maximizing the GMM likelihood, we can not only well position the Gaussian centers, but also adjust  the representations towards their associating cluster centers, as shown in Fig.  \ref{fig:GaussianDistributions} (b). In this way, we can enhance the compactness of the clusters.
We also explicitly maximize the distance between the Gaussian centers, which is achieved by iteratively update the Gaussian centers to make them far away from each other, as shown in Fig. \ref{fig:GaussianDistributions} (c).
By doing this, we can enhance the separability of the clusters. This can also implicitly enlarge the  inter-cluster distance between data sample representations.

Fundamentally, our contributions can be highlighted as: (i) we propose a new network structure for joint optimization of both deep representation learning and GMM-based modeling; (ii) the proposed framework can learn representations which are inter-cluster separable and intra-cluster compact; (iii) we model a deep representation of the whole data set as a GMM, and expect the data from the same cluster share a Gaussian component.

\section{Related Work}
\label{sec:related-work}


Clustering has been widely applied in many computer vision tasks. Popular clustering methods include k-means \cite{Lloyd1982Least}, GMM \cite{Bishop:2006:PRM:1162264}, non-negative matrix factorization\cite{Ding2010Convex}, and spectral clustering \cite{Zelnik2004Self-nips-2004}. Based on the low-level features, the clustering methods can perform well on a limited number of tasks  \cite{Aggarwal:2013:DCA:2535015}.

Since the popularity of deep learning, researchers tend to conduct clustering based on the deep features. Hinton and Salakhutdinov \cite{Hinton-science-2006} propose to train deep autoencoder networks and take the outputs of a bottleneck central layer as the representations. Schroff et al.  \cite{Schroff-facenet-cvpr} train FaceNet to extract features that can reveal the similarity between face images. 
Bruna and Mallat  \cite{Bruna2013Invariant} propose a wavelet scattering network to learn image representations that are stable to deformation and feasible for both classification and clustering.

The generic deep features can be applied in many different tasks and achieve better performance than the hand-crafted features  \cite{Hinton-science-2006,Schroff-facenet-cvpr,Bruna2013Invariant}. However, they are not necessarily optimal for the task of clustering.
To further improve the clustering performance, researchers  \cite{yangCVPR2016joint,yang_icml2017-pmlr-v70-yang17b,Xie-icml-2016} propose to integrate the representation learning with clustering. For joint optimization, Yang et al. \cite{yang_icml2017-pmlr-v70-yang17b} introduce an objective function consisting of three parts, i.e. dimension reduction, data reconstruction, and cluster structure regularization. Based on the idea of agglomerative clustering, Yang et al. \cite{yangCVPR2016joint} propose a recurrent framework for unsupervised clustering. By introducing an auxiliary distribution, Xie et al. \cite{Xie-icml-2016} propose another method for joint optimization.

While all the existing approaches for joint optimization have achieved certain level of success as reported in the literature, none of them directed the joint optimization towards improving the compactness and separability in clustering, which remains crucial for unsupervised deep learning among unlabeled data sets. To this end, we propose a new joint optimization approach for both representation learning and clustering, which simultaneously increases the separability and compactness for all the evolved clusters.

Significant research efforts have been reported to model the distribution of image representations, and achieve good performance in a variety of computer vision tasks, such as scene categorization  \cite{Nakayama2010Global} and image classification  \cite{Serra2015GOLD}. 
In our newly proposed approach, we model the distribution of data representations  (from many different clusters) with a GMM, and expect the data from the same cluster share a Gaussian component.

Modeling the features by a GMM has been studied in the research area of automatic speech recognition \cite{Sainath2012Auto,Deng2014Sequence}.
The work  \cite{Paulik2013Lattice} proposes a framework for bottleneck feature extraction. However, this work does not update the GMM parameter in the back-propagation procedure.
Based on the observation that log-linear mixture model (LMM) is equivalent to GMM  \cite{Heigold2010A}, the work  \cite{tuske2015Integrating} transforms GMM to LMM and implements it using popularly used neural network elements.
As reported \cite{Heigold2011Equivalence}, the soft-max layer in CNN is equivalent to a single Gaussian model with a globally pooled covariance matrix.
The work  \cite{Variani2015A} applied a joint optimization strategy of feature extraction and classification in the task of automatic speech recognition.  However, to the best of our knowledge, the joint optimization of CNN and GMM has not been studied in the area of unsupervised clustering.

\section{The Proposed Approach}
\label{sec:the-proposed-approach}

\subsection{GMM}

A Gaussian Mixture Model (GMM) expresses the probability as a wighted sum of a finite number of Gaussian component densities, as follows
\begin{equation}
\label{eq:GMM-function}
p(x|\lambda)=\sum_{k=1}^{m}{\omega_kg(x|\mu_k,\Sigma_k)}
\end{equation}
where  $ x \in R^d $ is a continuous-valued feature vector, $ m $ is the number of Gaussian components,  $ \omega_k (k=1,\cdots,m) $ are the mixture weights, and $ g(x|\mu_k,\Sigma_k) (k=1,\cdots,m)$ are the Gaussian densities.
The mixture weights satisfy the constraint that $ \Sigma_{k=1}^{m}\omega_k =1$.
Each Gaussian component density is a Gaussian function, i.e.
\begin{equation}
\label{Gaussian_component_calculate}
g(x|\mu_k,\Sigma_k)= \frac{1}{\sqrt{(2\pi)^{d}|\Sigma_k|}}exp\{- \frac{1}{2}(x-\mu_k)^{T}\Sigma_{k}^{-1}(x-\mu_k) \}
\end{equation}
with mean vector $ \mu_k $ and covariance matrix $ \Sigma_k $. 
For simplicity, let  $ \lambda $ denote a combinational group of the mean, covariance, and the mixture weight of the Gaussian components, i.e.
$ \lambda =\{\omega_k, \mu_k, \Sigma_k\}, k=1,\cdots, m $.



In order to estimate the parameter $ \lambda $, we normally maximize the GMM likelihood formulated as
\begin{equation}
\label{GMM_likeli_hood}
p(X|\lambda)=\prod\limits_{n = 1}^N p(x_n|\lambda)
\end{equation}
where $ X= \{ x_1, x_2, \cdots , x_N \}$ is a set of independent observations.
To solve this maximization problem, the expectation-maximization (EM) algorithm is widely applied, which improves the parameters iteratively with the following two steps. 

\textbf{Expectation}. Fix the parameter $ \lambda $, and calculate the posteriori probability of every sample belonging to each component. 


\textbf{Maximization}. With the above probability,  update the parameter of each Gaussian component to maximize the GMM likelihood (expressed in eq. \ref{GMM_likeli_hood} ).

%
%
%
%


\subsection{Representation Learning and GMM-based Modeling}
\label{subsec-representation-gmm-modeling}


Inspired by the fact that a proper representation learning procedure can significantly improve the clustering results  \cite{yang_icml2017-pmlr-v70-yang17b,law-icml2017-pmlr-v70-law17a}, we propose in this paper a new approach for integrated optimization of representation learning and clustering.

Regarding the representation learning, the distribution of the representations itself is another important factor to consider, in addition to the correlation between the representations and class labels.
The work  \cite{Stuhlsatz_feature_extraction_TNNLS2012} shows that we can learn a neural network that transforms arbitrary data distribution into a Gaussian distribution. The Gaussian distributed data representations are successfully applied in different computer vision tasks \cite{Nakayama2010Global,Serra2015GOLD,Wang2016RAID,Wang2017-cvpr-oral-ggdenet}.
Inspired by this, we propose to model the representations of the unlabeled data samples by a GMM and expect that the representations from the same cluster share a Gaussian component.

Let $ f_\theta(x) $ denote the representation of data sample $ x $ extracted by a convolutional neural network (with $ \theta $ as the parameter).
We model the distribution of $ f_\theta(x) $ as follows
\begin{equation}
f_\theta(x) \sim p(f_\theta(x)|\lambda)
\end{equation}
where the probability function $ p $ is a GMM formulated in Eq. \ref{eq:GMM-function} and $ \lambda $ denotes a combinational group of the GMM parameters, i.e.
$ \lambda =\{\omega_k, \mu_k, \Sigma_k\} $.

For joint optimization of both deep representation learning and GMM-based modeling, we maximize the following objective function
\begin{equation}
\label{eq:final_object_function}
O=log(P(f_\theta(X)|\lambda))+\eta S(\mu)
=log(\prod\limits_{n = 1}^N p(f_\theta(x_n)|\lambda))
+ \eta \sum_{k=1}^m \sum\limits_{j \in n(k)} d(\mu_k,\mu_j)
\end{equation}

Here, $ X=\{x_1,x_2,\cdots,x_N\} $ represents the whole data sample set and $ N $ is the number of data samples.
The parameter $ \eta $ is nonnegative to balance the two terms. Let $ n(k) $ denote the neighboring Gaussian components of the $ k $th component (measured by the distance $ d(x,y) $ between the centers of different components). While the first term in Eq. \ref{eq:final_object_function} calculates the log GMM  likelihood of the representations, the second term assesses the separability between different Gaussian components.

The first term in Eq. \ref{eq:final_object_function} has two sets of parameters, i.e. the CNN parameter $ \theta $ and the GMM parameter $ \lambda $.
With a fixed parameter $ \theta $ (and thus the representations of the data samples), a larger likelihood means the GMM can better model the distributions of the representations. With a fixed parameter  $ \lambda $, a larger likelihood means the sample representations are closer to their associated Gaussian centers. At the learning stage, we update the GMM parameter $ \lambda $  to better model the data representations, and update the CNN parameter $ \theta $ to adjust the data representations towards their associating centers (which can enhance the compactness of each Gaussian component).
Correspondingly, maximizing the first term in Eq. \ref{eq:final_object_function} guarantees: 1) that the GMM can well model the data representations; and 2) that each Gaussian component corresponds to a compact cluster in the data representation space.

By maximizing the second term in Eq. \ref{eq:final_object_function}, we can enhance the separability of the Gaussian components and thus improve the clustering performance. In addition to compactness, separability is another important criteria in clustering tasks. 
When we enlarge the distance between the Gaussian centers, we implicitly increase the distances between the data representations belonging to different components.
In addition, the introduction of the separability term also guarantees that the GMM model and the data representations are not trivial (i.e. all of data samples sharing the same representation).

To optimize our proposed framework for CNN-based representation learning, we update the parameters $ \theta $ and $ \lambda $ iteratively based on the the evaluations of data samples.
With a data sample $ x_n $ as the input, specifically, we introduce and maximize the following objective function
\begin{equation}
\label{eq:sample_based_object_function}
O(x_n|\lambda,\theta)=log( p(f_\theta(x_n)|\lambda))
+ \eta \sum_{k=1}^m \sum\limits_{j \in n(k)} \|\mu_k-\mu_j\|^2
\end{equation}
To speed up the parameter learning process, we restrict the covariance matrix to be diagonal, i.e. $ \Sigma=diag(\sigma_{1}^2,\sigma_{2}^2,\cdots,\sigma_{D}^2) $, where $ D $ is the dimensionality of the representations. Let $ y $ denote the representation of $ x $, i.e. $ y=f(x) $.
Out of the basic mathematical derivations, we achieve the following deviations as given in Eq. \ref{eq:deriveOfX}-\ref{eq:deriveOfomega} for the objective function regarding the parameters, where the index $1 \leq k \leq m $ corresponds to the Gaussian components, and $1 \leq d \leq D $ denotes the dimension of the representations (or the parameters), i.e. $ \mu_{kd} $ and $ y_d $
respectively denotes the $ d $th dimension of the mean vector and the representation.

\begin{equation}
\label{eq:deriveOfX}
\frac{\partial O(x|\lambda,\theta)}{\partial y}
=  
\sum_{k=1}^{m} {p(c_k|y) \Sigma_k^{-1}(\mu_{k}-y)}
\end{equation}

\begin{equation}
\label{eq:deriveOfMu}
\begin{aligned}
\frac{\partial O(x|\lambda,\theta)}{\partial \mu_{kd}} 
=
p(c_k|y) \frac{y_d -\mu_{kd}}{\sigma_{kd}^{2}}  + 2\eta \sum_{j\in n(k)} (\mu_{kd}-\mu_{jd})
\end{aligned}
\end{equation}

\begin{equation}
\label{eq:deriveOfsigma}
\frac{\partial O(x|\lambda,\theta)}{\partial \sigma_{kd}}
=
p(c_k|y) [\frac{(y_d-\mu_{kd})^2}{\sigma_{kd}^{2}}-1] 
\end{equation}

\begin{equation}
\label{eq:deriveOfomega}
\frac{\partial O(x|\lambda,\theta)}{\partial \omega_{k}}
=
p(c_k|y) -\omega_{k}
\end{equation}
where $ p(c_k|y)=\left. \omega_k g(y|\mu_{k},\Sigma_{k}) \middle/ p(y|\lambda) \right. $
denotes the probability that sample $ x_i $ belonging to the $ k $th Gaussian component.
In a backpropagation stage, we can use Eq. \ref{eq:deriveOfX} to update the parameter $ \theta $ and thus the representation $ y $. The parameters of the GMM are updated based on Eq. \ref{eq:deriveOfMu},  \ref{eq:deriveOfsigma}, and \ref{eq:deriveOfomega}.

%
%
%
%
%
%

\subsection{Network Structure}
\label{subsection-network-structure}

\begin{figure*}[t]
	\begin{center}
		\includegraphics[width=0.8\linewidth]{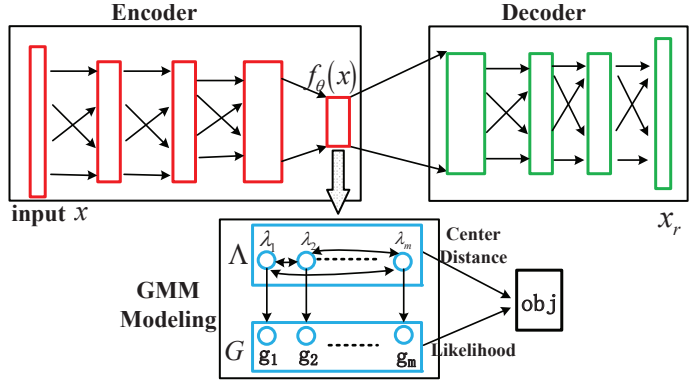}
	\end{center}
	\caption{The proposed network structure. The network consists of three components, i.e. the encoder, the decoder, and the representation modeling network.
	The representation modeling network models the data representations by a GMM.  Each node $ \lambda $ represents the weight, the mean and covariance matrix of a Gaussian component. The $ \Lambda $ layer is initialized based on the deep features $ f_\theta(x) $ produced by the encoder. Note that, the Gaussian components are mutually influenced by each other. }
	\label{fig:network}
\end{figure*}

To complete our proposed framework for integrated representation learning and clustering, we propose a network structure (as shown in Fig. \ref{fig:network}) consisting of three components, i.e. the encoder, the decoder, and the representation modeling network (RMN). Three steps are designed for its training.
Firstly, we train the encoder and the decoder by the data samples. Secondly, we initialize the RMN by a GMM that best captures the distribution of data representations produced by the encoder. Finally we jointly optimize the encoder and the RMN.

It is widely recognized that autoencoders  can learn representations that are semantically meaningful \cite{Hinton-science-2006}. This work trains an autoencoder and uses the encoder subnetwork to initialize the representation extraction network. We train the denosing autoencoder layer-by-layer. At the training stage, we first randomly corrupt the input data sample, and then use the denosing autoencoder to reconstruct the clean sample. The mathematical expression of a one layer denosing autoencoder is given as follows
\begin{equation}
\label{eq:dropout}
\widetilde{x}\sim q_D(\widetilde{x}|x)
\end{equation}
\begin{equation}
\label{eq:encoder-transform}
y=f_{\theta_1}(\widetilde{x}) =s(W_1\widetilde{x}+b_1)
\end{equation}
\begin{equation}
\label{eq:decoder-transform}
z=g_{\theta_1^{'}}({y}) =s(W_1^{'}y+b_1^{'})
\end{equation}
and the objective function is the squared distance between the input data sample and the reconstruction result, i.e.  $ ||x-z||^2 $. The function $ q_D $ in Eq. \ref{eq:dropout} denotes a stochastic mapping, i.e. randomly chooses a portion of data sample dimensions and set them to be $ 0 $. Let $ y $ be the representation extracted by the parameters $ W_1 $ and $ b_1 $, and the reconstruction result is denoted by $ z $ and the decoder parameters by $ W_1^{'} $ and $ b_1^{'} $. The autoencoder can be easily extended to multiple layers.


The RMN consists of two layers, one $ \Lambda $ layer corresponding to the parameters of the $ m $ Gaussian components and one $ G $ layer corresponding to the mixture of Gaussian components. 
In the network, a node $ \lambda_k =\{\mu_k,\Sigma_k,\omega_k\} $ denotes the parameters of the $ k $th Gaussian component. 
A node  $ g_k (1 \leq k \leq m)$ denotes the value of the $ k $th Gaussian component evaluated by Eq. \ref{Gaussian_component_calculate}. Both the two layers take the data representations produced by the encoder as input. 
For the $ \Lambda $ layer, the data representations are used to initialized the parameters. For the $ G $ layer, the data representations are used to evaluated the likelihood.
Note that, due to the second term in Eq. \ref{eq:final_object_function}, the centers of the Gaussian components are mutually influenced by each other. 
Each of these two layers contributes to a term of Eq. \ref{eq:sample_based_object_function}, i.e. the $ \Lambda $ layer corresponds to the separability between clusters and the $ G $ layer corresponds to the compactness of clusters in addition to the feasibility of the GMM. Thus, both of these two layers contribute to the objective function.

\section{Experiments}
\label{sec:experiments}

\subsection{Dataset}

To evaluate the proposed method, we conduct experiments on six datasets, i.e. MNIST, USPS, COIL20, COIL100, STL-10, and Reuters. 


%

The MNIST dataset \cite{LeCun-Ieee-1998-mnist} is one of the most popular image datasets. It consists of 60,000 training samples and 10,000 testing samples from 10 classes (from $ 0 $ to $ 9 $). Each image in the dataset represents a handwritten digit.
The images are centered with size of $ 28 \times 28 $.

USPS\footnote{https://cs.nyu.edu/~roweis/data.html} is a another handwritten digits dataset produced by the USPS postal service. In total, this data set contains 11,000 samples belonging to the 10 different classes, where the image size is $ 16 \times 16 $.

COIL20 \cite{Nene-1996-coil20} and COIL100 \cite{Nene-1996-coil100} are two datasets built by Columbia University, which respectively contain  $ 1,440 $ gray images of $ 20 $ objects and $ 7,200 $ color images of $ 100 $ objects. The images are captured under different views.

The STL-10 dataset \cite{stl10-pmlr-v15-coates11a} consists of images from $ 10 $ different classes: \textit{airplane, bird, car, cat, deer, dog, horse, monkey, ship, truck}. Each class has  $ 1300 $ labeled images. In addition, there are also $ 100,000 $ unlabeled images. Note that, the unlabeled image set contains images not belonging to the above 10 classes.  The size of the images is $ 96 \times 96 $. While we use the labeled data to test our method, the unlabeled data are also used to train the autoencoder network. Following \cite{Doersch:2012:MPL:2185520.2185597}, we calculate the $ 8\times8 $ color map and the HOG features of each image, and take the concatenation of them as the input.

The REUTERS dataset \cite{reuters-dataset-Lewis:2004:RNB:1005332.1005345} contains $ 804,414 $ documents  from $ 103 $ different topics. 
We use a subset of this dataset, which contains $ 365,968 $ documents from $ 20 $ different topics. 
As in \cite{nitish-JMLR:v15:srivastava14a}, we use the tf-idf features of those most frequently used words to represent the documents.

%

\subsection{Implementation Details}
In the autodecoder, we adopt the widely used rectified linear units (ReLUs), 
except the layer where the data samples are reconstructed and the layer where the representations are produced
\cite{vincent-jmlr2010}.
Inspired by the work \cite{pmlr-v5-maaten09a}, the encoder consists of $ 3 $ fully connected layers (excluding the input layer), and their numbers of nodes are respectively $ 500 $, $ 500 $, and $ 2000 $. 
The representations of the input is extracted by a fully connected layer with $ 10 $ nodes. The
decoder is a mirrored version of the encoder.
After the encoder and decoder are trained layer-by-layer in a greedy manner, we concatenate them together and fine-tune the whole network. We then use the encoder subnetwork to produce the representations of data samples. In other words, the encoder can be considered as the initialization of the representation learning network.


The RMN is initialized based on the distribution of data sample representations.
Specifically, we first extract the representations of all the data samples using the encoder, and then learn the initialized GMM based on these representations. To maintain a fair comparison with the benchmark methods, the number of the Gaussian components in the GMM is equal to the number of clusters in the dataset. The parameters of this GMM are adopted to initialize the RMN, and each node $ \lambda_i $ corresponds to the three parameters corresponding to a Gaussian component, i.e. the coefficient, the mean and the covariance. After that, we adopt the  SGD (stochastic gradient descent) method to jointly optimize the encoder and RMN.  We set the base learning rate to be $ 0.01 $ and take the step policy to update the learning rate. 
We set the cardinality  $ |n(k)| $ (in Eq. \ref{eq:final_object_function}) to be half of the number of clusters. In this way, the center of one cluster is influenced by half of the remaining clusters that are nearby. 
For the parameter $ \eta $ in Eq. \ref{eq:final_object_function}, we choose the best one from $ \{0.1,0.01,0.001,0.0001\} $.

\subsection{Benchmarks}

We compare the proposed method with a number of unsupervised clustering methods. 
Firstly, we take K-means \cite{Lloyd1982Least} and GMM \cite{Bishop:2006:PRM:1162264} as the baseline benchmarks. They can either take the low-level feature, or the deep autoencoder feature (AEF) as input.

Secondly, we compare our method with  two agglomerative methods, i.e. agglomerative clustering (AC-GDL) \cite{Zhang-Graph-degree:2012:GDL:2402940.2402972} and agglomerative clustering via path integral (AC-PIC) \cite{Zhang-agglomerative-PR2013-DBLP:journals/pr/0010ZW13}.  We also take  two subspace-based clustering methods as benchmarks, including large-scale spectral clustering (SC-LS) \cite{Chen-Large-Scale-AAAI:2011:LSS:2900423.2900472} and NMF with deep model (NMF-D) \cite{Trigeorgis-Adeep-icml:2014:DSM:3044805.3045081}.

Thirdly, we compare our method with four other benchmarks, which also jointly optimize the representation learning and clustering, i.e. DEC \cite{Xie-icml-2016}, Joint unsupervised learning (JULE) \cite{yangCVPR2016joint}, DCN \cite{yang_icml2017-pmlr-v70-yang17b}, and DEPICT \cite{Dizaji2017DeepCV}.

Finally, we  set the parameter $ \eta $ in Eq. \ref{eq:final_object_function} to be zero and produce another benchmark (denoted as DeepGMM in this paper) to assess the effectiveness of the second term in Eq. \ref{eq:final_object_function}, in terms of improving the separability between different Gaussian components. The only difference from the proposed method is that DeepGMM does not explicitly enlarge the distances between the Gaussian centers.

\begin{table*}[!htb]
	\centering
	\caption{The accuracy of the proposed method and the benchmarks on six datasets }
	\begin{tabular}{l|c|c|c|c|c|c}
		\hline
		{Dataset} & MNIST  & USPS  & COIL 20  & COIL 100  & STL-10  & REUTERS   \\
		\hline \hline
		K-Means \cite{Lloyd1982Least}  &  53.5\% &  46.0\% &  48.3\%& 51.4\% &28.4\% & 32.3\%  \\		
		GMM  \cite{Bishop:2006:PRM:1162264} &  47.6\% &  64.2\%  & 54.3\%& 67.5\% &20.3\% & 26.6\%  \\		
		
		AEF+KM   &  80.0\% &  64.3\% &  54.1\%& 67.5\% &29.7\% & 35.8\% \\		
		
		AEF+GMM  &  64.1\% &  71.3\% &  69.8\%& 73.8\% &22.2\% & 31.6\%  \\
		\hline
		\hline
		AC-GDL \cite{Zhang-Graph-degree:2012:GDL:2402940.2402972} &  11.3\% &  {86.7\%} &  76.5\%& 80.5\% & 26.8\% & 36.1\% \\		
		
		AC-PIC \cite{Zhang-agglomerative-PR2013-DBLP:journals/pr/0010ZW13} &  11.5\% &  85.5\% &  70.3\%& {84.6\%} & 24.1\% & 29.4\%  \\				
		SC-LS \cite{Chen-Large-Scale-AAAI:2011:LSS:2900423.2900472}  &  71.4\% &  65.9\% & {76.4\%}& {82.6\%} & 20.4\% & 37.2\%  \\
		NMF-D \cite{Trigeorgis-Adeep-icml:2014:DSM:3044805.3045081} &  17.5\% &  38.2\% &  64.3\%& 70.2\% &30.6\% & {39.5\%}  \\				
		\hline
		\hline
		DEC \cite{Xie-icml-2016} &  84.4\% &  61.9\% &  83.6\%& 75.5\% & {35.9\%} & 14.0\%  \\				
		
		JULE \cite{yangCVPR2016joint}  &  {90.6\%} &  {91.4\%} &  80.0\%& 77.4\% & 17.7\% & 38.1\%  \\				
		
		DCN \cite{yang_icml2017-pmlr-v70-yang17b} &  83.0\% & 77.8\% &  76.5\%& 69.7\% & {34.1\%} &  {47.0\%} \\				
		DEPICT  \cite{Dizaji2017DeepCV} &   91.2\% &   91.4\% &  81.3\% & 76.4\%& 32.8\% & 29.9\%  \\						
		DeepGMM  &  72.5\% &  65.4\% &  72.3\%& 52.9\% &27.6\% & 51.3\%  \\		
		\hline
		\hline
		\textbf{The proposed}  &   \textbf{93.9\%} &   \textbf{94.7\%} &   \textbf{88.5\%}&  \textbf{85.1\% }&  \textbf{36.3\%} &  \textbf{56.9\%}  \\				
		\hline
	\end{tabular} 
	\label{tab:dataset-accuracy}
\end{table*}

\begin{table*}[!htb]
	\centering
	\caption{The NMI of the proposed method and the benchmarks on six datasets}
	\begin{tabular}{l|c|c|c|c|c|c}
		\hline 
		{Dataset} & MNIST & USPS  & COIL 20 & COIL 100  & STL-10  & REUTERS  \\
		\hline \hline
		K-Means \cite{Lloyd1982Least}  &  0.50 &  0.45  & 0.74& 0.78 &0.25 & 0.17 \\				
		GMM  \cite{Bishop:2006:PRM:1162264} & 0.46  &  0.63 & 0.51& 0.75 &0.16 & 0.38  \\		
		AEF+KM &  0.73 &  0.59 & 0.77& 0.82 &0.26 &  {0.41} \\		
		
		AEF+GMM  &   0.59&  0.68 &  0.60 & 0.81 & 0.20 & {0.48}  \\
		\hline
		\hline
		AC-GDL \cite{Zhang-Graph-degree:2012:GDL:2402940.2402972} &  0.12 &  0.82  & 0.80& {0.78} &0.21 & 0.38 \\		
		
		AC-PIC \cite{Zhang-agglomerative-PR2013-DBLP:journals/pr/0010ZW13} &  0.12 &  0.84  &  {0.79}&  {0.81} &0.18 & 0.27 \\				
		SC-LS \cite{Chen-Large-Scale-AAAI:2011:LSS:2900423.2900472}  &  0.71 &  0.68  &  {0.77} & 0.83 &0.16 &0.34  \\				
		NMF-D \cite{Trigeorgis-Adeep-icml:2014:DSM:3044805.3045081} &  0.15 &  0.29  & 0.69& 0.72 &0.24 & 0.31  \\				
		\hline
		\hline
		DEC \cite{Xie-icml-2016} &  0.80 &  0.58 &  {0.84} & {0.79} &  {0.31} & 0.28  \\				
		
		JULE \cite{yangCVPR2016joint}  &   {0.87} &  {0.88}  & 0.85& 0.83 &0.14 &0.36  \\				
		
		DCN \cite{yang_icml2017-pmlr-v70-yang17b} &  0.81 &  0.85 & 0.79 & 0.74 &  {0.30} &  {0.51}  \\				
		DEPICT  \cite{Dizaji2017DeepCV} &   0.87 &   0.88 & 0.84 & 0.84& 0.36 & 0.48 \\				
		DeepGMM  &  0.64 &  0.51 &  0.74 & 0.51 &0.21 & 0.49 \\
		\hline
		\hline				
		\textbf{The proposed}  &   \textbf{0.87} &   \textbf{0.92} &  \textbf{0.89} &  \textbf{0.90} & \textbf{0.34} &  \textbf{0.56}  \\				
		\hline
	\end{tabular} 
	\label{tab:dataset-nmi}
\end{table*}

%
%
%
%
%
%
%
%
%

\subsection{Performance}

Three popular standard metrics in evaluating the clustering algorithms are adopted, which include clustering accuracy (ACC) \cite{Xu2003Document}, normalized mutual information (NMI) \cite{cai-document-tkde},  Calinski-Harabaz score (CH) \cite{Cali1974A}.

The ACC is defined as 
$ ACC=\frac{1}{N} \sum_{i=1}^{N}\delta(l_i,map(r_i))$,
where $ N $ is the total number of data samples, $ l_i $ denotes the ground truth cluster label, and $ \delta (x,y) $ is the delta function which equals $ 1 $ iff its two parameters are the same.


Let $ C $ and $ R $ respectively denote the clustering results and the groundtruth clusters, the NMI  is defined as $NMI(C,R)=\left. MI(C,R) \middle/ max(H(R),H(C)) \right.$ where $ MI(C,R) $ is the mutual information between $ C $ and $ R $, and $ H(R) $ and $ H(C) $  are the entropies.

Let $ k $ denote the number of clusters, the CH score is defined based on the between-clusters dispersion mean matrix $ B_k $ and within-cluster dispersion matrix $ s(k)=\frac{tr(B_k)}{tr(W_k)}\times\frac{N-k}{k-1}$.
The CH score is higher when the resulting clusters are compact and well separated.

Tab. \ref{tab:dataset-accuracy} and Tab. \ref{tab:dataset-nmi} respectively list the 
accuracy and NMI of the proposed methods in comparison with the benchmarks. 
As seen, the
proposed method achieves the highest ACC and NMI on all of five image datasets and one text dataset, which indicates that the proposed method can learn feasible deep representations for the clustering task in different applications. Generally speaking, the methods involving deep representations perform better than the ones with low-level features.  Taking the popular k-means and GMM as the examples,  we can always improve the accuracy and NMI by learning deep representations in our experiments. Thus, it is necessary to proposed unsupervised deep representation learning methods for  clustering.

\begin{figure*}[t]
	\begin{center}
		\includegraphics[width=0.9\linewidth]{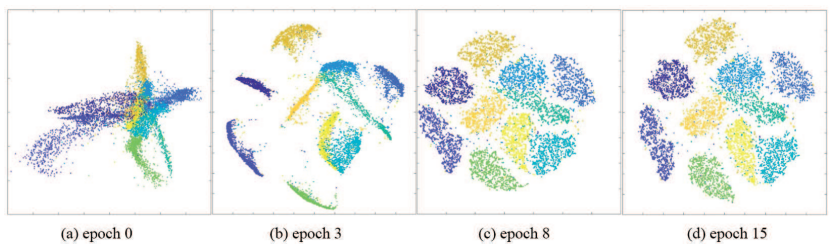}
	\end{center}
	\caption{The clusters of MNIST in different epochs. While the initial representations are mixture together, they gradually evolve into separable clusters. In addition, the compactness of the clusters is also gradually improved.}
	\label{fig:epochs}
\end{figure*}

The results in these two tables also illustrate that
the proposed method always performs better than DeepGMM, which validates that the introduction of the second term in Eq. \ref{eq:final_object_function} can indeed improve the separability of the data representations. In other words, by maximizing the distance between the Gaussian centers, we enlarge the distances of representations belonging to different classes. In Fig. \ref{fig:epochs}, we visualize the data representations of a MNIST subset (with $ 10,000 $ images) in different epochs using  t-distributed stochastic neighbor embedding (t-SNE) \cite{pmlr-v5-maaten09a}. We can clearly see that the clusters are gradually become more compact and more separable. To explicitly assess the compactness and separability of the resulting clusters, we list the CH-score of different deep learning methods in Tab. \ref{tab:dataset-ch-score}. As seen, the proposed method achieves the highest CH-score, indicating the resulting clusters are more compact and more separable.

\begin{figure}[!htb]
	\centering
	\hfill
	\subfigure[ MNIST dataset]{\includegraphics[width=.49\linewidth]{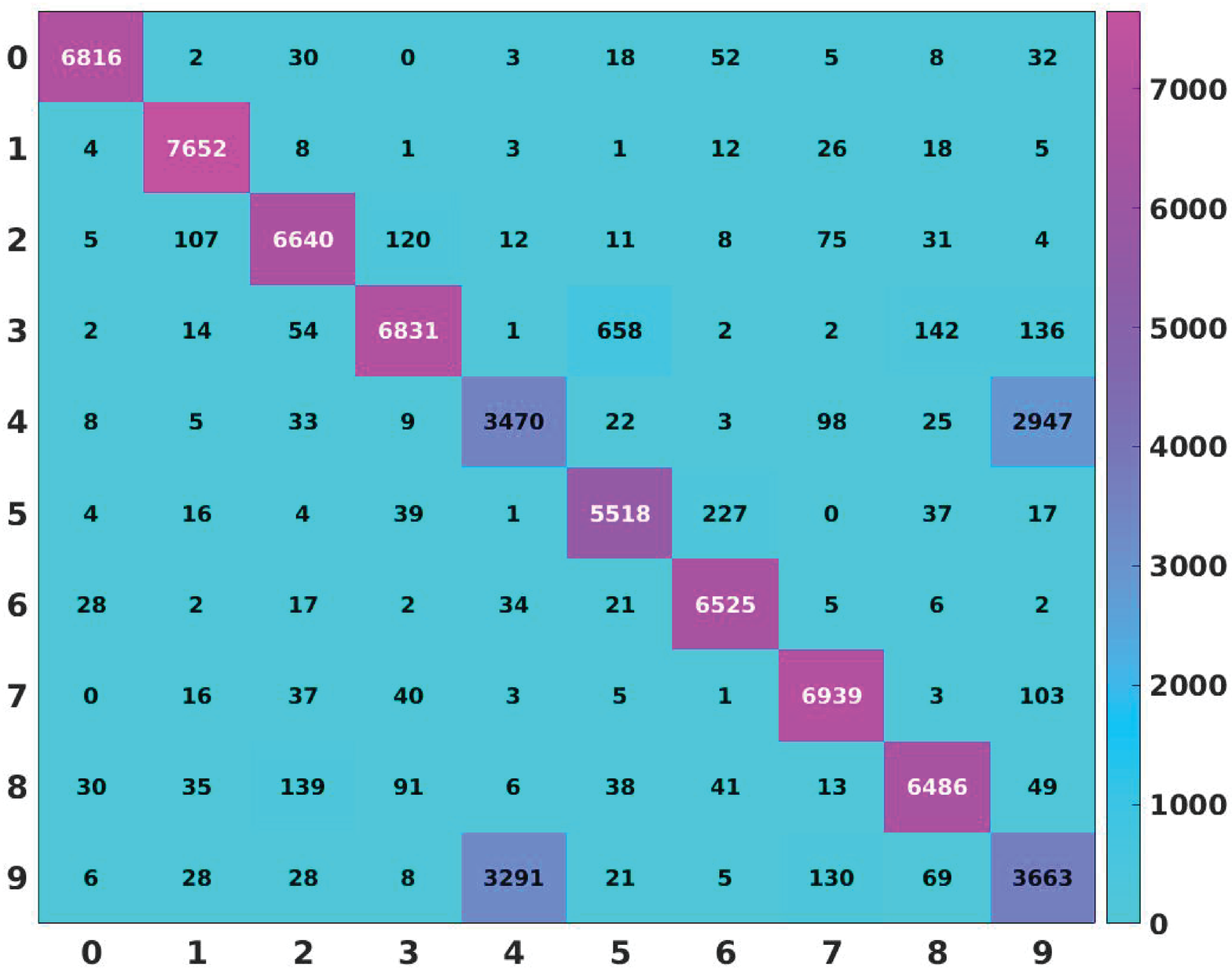}}
	\subfigure[STL-10 dataset]{\includegraphics[width=.49\linewidth]{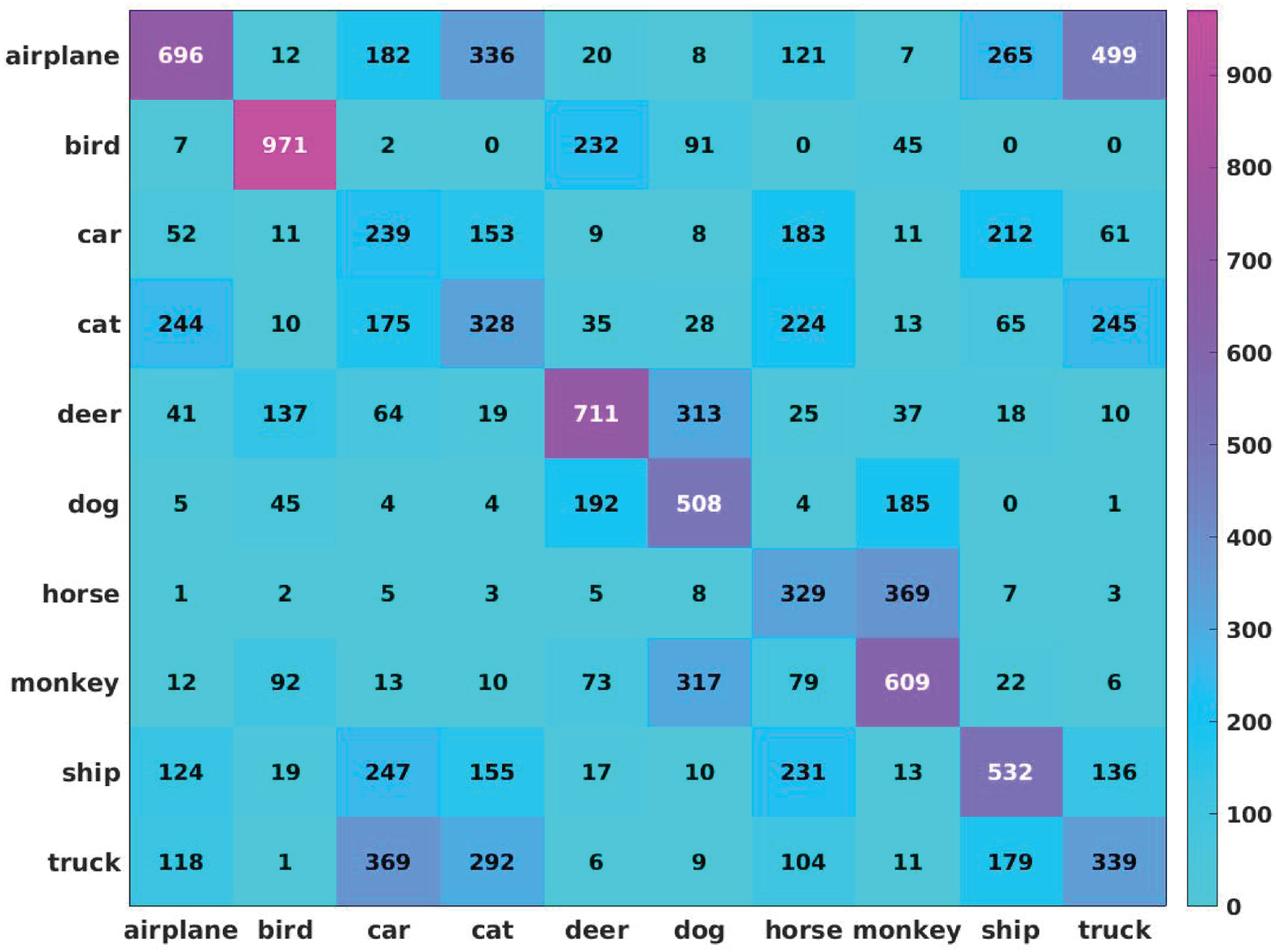}}
	\hfill
	\caption{The confusion matrices of MNIST dataset and STL-10 dataset. Each row corresponds to a resulting cluster and each column corresponds to a groundtruth cluster. }
	\label{fig:confusion-matrix}
\end{figure}

The proposed method significantly outperforms the benchmark AE+GMM, which indicates that our proposed  joint optimization can produce more clustering-friendly representations. While autoencoder can extract semantic meaningful representations, the proposed method can significantly enhance their discriminant ability.

\begin{table*}[!htb]
	\centering
	\caption{The CH-score of different deep learning methods on six datasets}
	\begin{tabular}{l|c|c|c|c|c|c}
		\hline 
		{Dataset} & MNIST & USPS  & COIL 20 & COIL 100  & STL-10  & REUTERS  \\
		\hline \hline
		DEC \cite{Xie-icml-2016} & 2172  &  274  &   72  &   64  &   41  &   88  \\				
		
		JULE \cite{yangCVPR2016joint}  &   1977  &  228  &   68  &   57  &   82  &  112  \\				
		
		DCN \cite{yang_icml2017-pmlr-v70-yang17b} & 2270  &  304  &   57  &   48  &   48  &   66 \\				
		
		DeepGMM  &  1684  &  199  &   64  &   42  &   64  &   55 \\
		\hline
		\hline				
		\textbf{The proposed}  & 2441  &  327  &   86  &   79  &  119  &  150 \\				
		\hline
	\end{tabular} 
	\label{tab:dataset-ch-score}
\end{table*}

\subsection{Discussion}

This subsection discusses the experimental results on MNIST and STL-10. Fig. \ref{fig:confusion-matrix} shows the confusion matrices of the proposed method on MNIST and STL-10.
As seen in Fig. \ref{fig:confusion-matrix} (a), we can know that the difficulty of the MNIST dataset mainly lies in the separability of $ 4 $ and $ 9 $ from each other. For the STL-10 dataset, on the other hand,  we can achieve relatively better performance on the clusters whose background and pose do not change significantly, such as \textit{airplane} and \textit{bird}.

\begin{figure}[htb]
	\hfill
	\subfigure[Images in MNIST that are far from their associating  centers]{\includegraphics[width=0.90\linewidth]{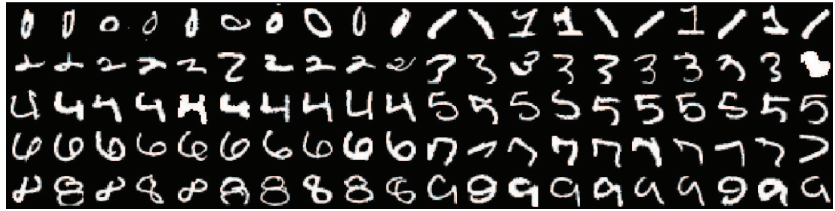}}
	\hfill
	\vfill
	\hfill
	\subfigure[Images in STL-10 that are far away from their associating centers]{\includegraphics[width=0.90\linewidth]{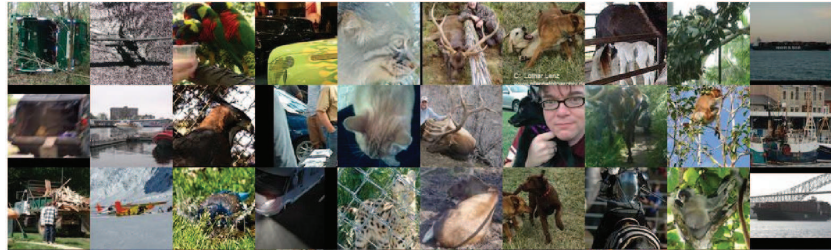}}
	\hfill
	\caption{The MNIST and STL-10 images which are far away from their associating centers}
	\label{fig:good_bad_examples}
\end{figure}

To identify the difficulty examples from the easier ones, we visualize the data samples which  are far away from the centers (in Fig. \ref{fig:good_bad_examples} for MNIST and   STL-10). 
For  MNIST, the center samples are the ones which are similar to the standard written characters. In other words, if the digit is well written, we can easily categorize it into the right cluster. On the contrary, the images which are far from their associating centers are not well written (if judged by common sense), as shown in Fig. \ref{fig:good_bad_examples} (a). Most of them are visually distorted. For some images, it is even difficult for human beings to recognize it correctly.

The complex background heavily affects the clustering performances for the STL-10 dataset. In addition, the pose of the foreground is another important factor for correct clustering. If the target foreground is well posed, its whole body is visible and can be easily clustered. Take the  cluster of \textit{car} as an example, the difficult images only contain a small portion of a car due to heavily side view capture or occlusion, as shown in Fig. \ref{fig:good_bad_examples} b). In these cases, it is difficult for our method to identify their similarities with the cars which are well posed. In our examples, if an image is captured from a side view, its representation is far from the corresponding cluster center   and thus are more likely  to be mis-clustered.

We vary the parameter   $ \eta $ to show the robustness of our method on MNIST. With $ \eta =10^{-1}, 10^{-2}, 10^{-3} $, and $ 10^{-4} $, the average accuracies of $ 10 $ times running are $ 90.3\%\pm 3.9\% $, $ 93.6\%\pm2.3\% $, $ 81.6\%\pm 5.6\% $, and $ 72.5\%\pm4.8\% $, respectively. The average NMIs are $ 0.84\pm0.02 $, $ 0.86\pm0.03 $, $ 0.80\pm0.04 $, and $0.68\pm0.05$, respectively.

\section{Conclusion}
\label{sec:conclusion}
In this paper, we have proposed and described an unsupervised deep learning framework by integrating deep representation with GMM-based modeling and joint optimization of representation learning and clustering. The deep representation learning procedure not only optimizes the compactness of each cluster corresponding to an individual Gaussian component inside the GMM, but also optimizes the separability across different clusters. As a result, the proposed network structure as shown in Fig. 2 can jointly optimize these two learning targets, and especially learn representations which are catered for the task of clustering. In addition, the optimization process simultaneously minimizes the distance between the representations and their associating centers, and maximizes the distances across different Gaussian centers. In this way, our proposed achieves the advantage that not only the compactness within individual clusters is improved, but also the separability across different clusters is enhanced, leading to significant improvements over the compared existing benchmarks.

\section*{Acknowledgment}
The authors wish to acknowledge the financial support from: (i) Natural Science Foundation China (NSFC) under the Grant No. 61620106008; (ii) Natural Science Foundation China (NSFC) under the Grant No. 61802266; and (iii) Shenzhen Commission for Scientific Research \& Innovations under the Grant No. JCYJ20160226191842793.

%
%
%

\bibliographystyle{splncs04}
\bibliography{accv-0190-bib}

\end{document}